\documentclass[lettersize,journal]{IEEEtran}
\usepackage{amsmath,amsfonts}
\usepackage{algorithmic}
\usepackage{algorithm}
\usepackage{array}
\usepackage[caption=false,font=normalsize,labelfont=sf,textfont=sf]{subfig}
\usepackage{textcomp}
\usepackage{stfloats}
\usepackage{url}
\usepackage{verbatim}
\usepackage{graphicx}
\usepackage{cite}
\usepackage{color}
\usepackage{xcolor}
\begin{document}

\title{Detecting Content Rating Violations in Android Applications: A Vision-Language Approach}

\author{D. Denipitiyage, B. Silva, S. Seneviratne, A. Seneviratne, S. Chawla
\thanks{D. Denipitiyage, B. Silva, and S. Seneviratne are with the School of Computer science, University of Sydney, Australia (e-mail: dden5444@uni.sydney.edu.au; bpin9254@uni.sydney.edu.au; suranga.seneviratne@sydney.edu.au)}
\thanks{A. Seneviratne is with the University of New South Wales (UNSW), Sydney, Australia (e-mail: a.seneviratne@unsw.edu.au) }
\thanks{S. Chawla is with the Qatar Computing Research Institute, Hamad Bin Khalifa University (HBKU) (e-mail: schawla@hbku.edu.qa)}}

\maketitle

\begin{abstract}
Despite regulatory efforts to establish reliable content-rating guidelines for mobile apps, the process of assigning content ratings in the Google Play Store remains self-regulated by the app developers. There is no straightforward method of verifying developer-assigned content ratings manually due to the overwhelming scale or automatically due to the challenging problem of interpreting textual and visual data and correlating them with content ratings. We propose and evaluate a vision-language approach to predict the content ratings of mobile game applications and detect content rating violations, using a dataset of metadata of popular Android games. 
Our method achieves $\sim$6\% better relative accuracy compared to the state-of-the-art CLIP-fine-tuned model in a multi-modal setting. Applying our classifier in the wild, we detected more than 70 possible cases of content rating violations, including nine instances with the `Teacher Approved' badge. Additionally, our findings indicate that 34.5\% of the apps identified by our classifier as violating content ratings were removed from the Play Store. In contrast, the removal rate for correctly classified apps was only 27\%. This discrepancy highlights the practical effectiveness of our classifier in identifying apps that are likely to be removed based on user complaints.
\end{abstract}


\makeatletter
\def\ps@IEEEtitlepagestyle{
  \def\@oddfoot{\mycopyrightnotice}
  \def\@evenfoot{}
}

\def\mycopyrightnotice{
  {\footnotesize
  \begin{minipage}{\textwidth}
  \centering
  This work has been submitted to the IEEE for possible publication. Copyright~\copyright~ may be transferred without notice, after which this version may no longer be accessible.
  \end{minipage}
  }
}

\section{Introduction}
\label{Sec:introduction_2}

In recent years, our reliance on mobile services has surged, whether for entertainment, communication, shopping, travel, or finance. According to recent reports, 60.42\% of the world's population owns a smart device~\cite{ashtuener2024}. One implication of this trend is that children of all ages are using smart devices and apps more than ever.

In 2013, a survey revealed that over 75\% of children under 8 years old were using mobile phones~\cite{liu2016identifying} and in 2019, 69\% of teens owned a smartphone~\cite{2019smartphone}. This widespread dependence among young children poses a significant social challenge whether they are being provided with age-appropriate content, usually defined by content ratings. For example, the Google Play Store ratings in the US and Canada are maintained by the Entertainment Software Rating Board (ESRB), with rating categories \emph{Everyone, Everyone 10+, Teen, Mature} and \emph{Adult only}, whereas in Australia, games adhere to local content maturity ratings issued by the Australian Classification Board (ACB), including G, PG, M, MA15+, and R18+. 
Similarly, rating systems are employed in different regions, such as Pan-European Game Information (PEGI) in Europe. These guidelines use six different content descriptors: themes, violence, sex, language, drug use (substances) and nudity and asses the impact depends on the frequency, intensity.

To assist users, especially parents, in finding suitable apps, the Google Play Store enforces strict developer policies. As a result, each app page displays critical app information, such as download counts, requested permissions, content rating, developer names, and user reviews, allowing users/ parents to understand the app before downloading.
% Google Play has several developer policies related to app content~\cite{google_developer_policy}.
Additionally,  all apps undergo automated inspection and vetting procedures before being published. Furthermore, in 2020, Play Store introduced ``Teacher Approved'' apps, which are published after being rated by teachers and specialists. They take into consideration factors such as design, age appropriateness, and appeal when rating an app~\cite{teacherAApps}. Despite such initiatives, Luo et al.~\cite{luo2020automatic} identified that 40\% of apps contained inappropriate content among 70 children's apps in 2020. Moreover, children reportedly spent 27\% more screen time for online video platforms, 120\% more for TikTok in 2023 compared to 2022~\cite{anualdatareport2023} despite those apps being rated for ages 13 and up. Furthermore, by analysing 20,000 apps in Google Play Store, Sun et al.~\cite{sun2023not} claimed that 19.25\% of apps have inconsistent content ratings across different protection authorities around the world, thus making them un-generalisable.

One reason why such content rating violations and discrepancies are possible among apps, especially in the Google Play Store, is its loosely regulated nature. The Google Play Store relies on an app developer's completed questionnaire and self-reported information to automatically determine the content rating as disclosed by developers~\cite{content_ratin_q_geographical}. In a profit-driven app ecosystem with over 3.6 million apps in the Google Play Store~\cite{avada}, where app engagements matter significantly, especially those from young children, it can not always be expected that all developers will play fair. Furthermore, Google employs different content rating systems based on geographical locations~\cite{content_ratin_q_geographical}, and there are no clear boundaries among the categories, to the extent that an average smartphone user can easily get overwhelmed. On the other hand, the app vetting process by the Apple App Store is manual~\cite{app_review_apple}, and as a result, it is most likely to have correct content ratings for apps; however, the downside is that getting apps published in App Store takes time.

As such, there is a stronger need to develop methods to automatically assign correct content ratings for given apps. This requirement is further exacerbated by the fact that regulatory bodies, such as a country's e-safety commission, do not have the necessary means to identify apps that violate the country's content rating guidelines unless end users complain about specific apps. Currently, such studies by regulators are mostly carried out manually or semi-manually. For example, in 2012 the FTC reviewed 200 apps, mostly through manual processes~\cite{federal2012mobile}.

To this end, in this paper, we propose a vision-language approach based on self-supervised learning and supervised contrastive learning that allows us to identify content rating violations in app markets. Our approach is based on the intuition that multi-modality is important in this problem (i.e., considering both app descriptions and images such as icons and screenshots, commonly known as app creatives). Within creatives, it is crucial to consider both the content and style of these images. The style information is effective in identifying the target demographic of an app, as apps designed for children tend to have cartoon effects and tactile textures like glitter. More specifically, we make the following contributions:

%Content rating violations can be intentional or unintentional; for example, a developer can intentionally assign an incorrect content rating to an app to increase the app's engagement and revenue. Equally, a developer can unknowingly assign an incorrect rating, making an unsuitable app available for underage users. Either way, automatically verifying whether an app has the correct content rating is essential. 

\begin{itemize}
    \item We propose a vision-language approach using trained content, style, and text encoders, along with a cross-attention module, to predict mobile app content ratings from descriptions and creatives.
    \item  Using a real-world dataset, we show that our approach achieves 5.9\% and 5.8\% relative improvements in accuracy over state-of-the-art CLIP and CLIP fine-tuned models.
    \item Upon evaluating the test dataset, our model identified 71 apps ($\sim$17\% of the total verified) with potential content rating malpractices in the Google Play Store and 32 apps subtly attracting an unsuitable audience. This includes nine `Teacher Approved' apps, which Google Play claims to verify manually.

    \item We conduct extensive experiments on nearly 16,000 apps to validate the effectiveness of our model. The results show that 45.7\% of identified malpractices and 39.1\% of identified disguises were removed by the Play Store after nine months of our initial crawl.
\end{itemize}

\section{Related Work}
\label{Sec:related_work}

\textbf{Automatic app maturity ratings}: The evaluation of mobile apps often involves various perspectives. In particular, identifying mobile app development is consistent with what is stated in the privacy policy concerning online advertising and tracking ~\cite{nguyen2022freely, nguyen2021measuring}, aiding developers in crafting child-friendly apps concerning both content and privacy aspects~\cite{hu2015protectingcikm, liccardi2014can}. However, fewer studies aimed at mobile app maturity rating. Therefore, there is growing concern regarding inappropriate content and maturity ratings in mobile apps, which are linked to privacy concerns. Early work by Chen et al.~\cite{chen2013isthisapp} proposed Automatic Label of Maturity ratings (ALM), a text-mining-based semi-supervised algorithm that uses app descriptions and user reviews to determine maturity ratings. The authors used the content rating from Apple App Store as the reference standard for a given app. However, this method uses keyword matching while ignoring semantic analysis. Using a similar approach for ground truth establishment, Hu et al.~\cite{hu2015protectingcikm} proposed a text feature-based SVM classifier for content rating prediction with an online training element. The previous two methods solely depend on text features despite having access to other modalities. Liu et al.~\cite{liu2016identifying} and Chenyu et al.~\cite{zhou2022automatic} extended these works by incorporating image and APK features to identify children’s apps. However, features were limited to extracting text using OCR software, colour distribution of the icon and screenshots, and permissions and APIs. More recently, Sun et al.~\cite{sun2023not} identified discrepancies in content ratings of the same app in different geographic regions by defining rating system mappings between geographical regions. However, this research focuses on single modalities or multiple modalities but treats them independently. \\ 
% \vspace{-3mm}

\noindent\textbf{Vision-Language (VL) models}:  Early image-based contrastive representations have made advancements, nearly achieving the performance levels seen in supervised baselines across various downstream tasks such as image classification and retrieval~\cite{chen2020simple, zbontar2021barlow}. Driven by the success of contrastive learning in intra-modal tasks, there has been a growing interest in developing multi-modal objectives (e.g., Vision-Language), enabling the model to comprehend and exploit cross-modal associations.
Pioneering works such as CLIP~\cite{clip} and ALIGN~\cite{align} bridged the gap between the vision and language modalities by learning language and vision encoders jointly with a symmetric cross-entropy loss which is an adaptation of InfoNCE loss~\cite{oord2018representation} for cross-model pairs. CLIP optimises the cosine similarity between text and image embeddings, while ALIGN employs a similar contrastive learning setting with noisy training data. Zhai et al.~\cite{LiT} tuned the text encoder using image-text pairs while keeping the image encoder frozen. The rich embeddings that these methods learn are later adapted to various application domains such as video-text retrieval~\cite{fang2021clip2video, portillo2021straightforward}, image generation~\cite{nichol2021glide}, and visual assistance~\cite{massiceti2023explaining}. 
However, \cite{agarwal2021evaluating, luccioni2024stable} point out the challenges in adapting Large Multi-modal Models (LMMs) for different domains when the downstream task deviates from the originally pre-trained task. To the best of our understanding, ours is the first work to leverage the advances in VL-language models to detect content compliance malpractices specific to mobile apps.

\begin{figure*}[ht]
    \centering
    \includegraphics[width=0.97\linewidth]{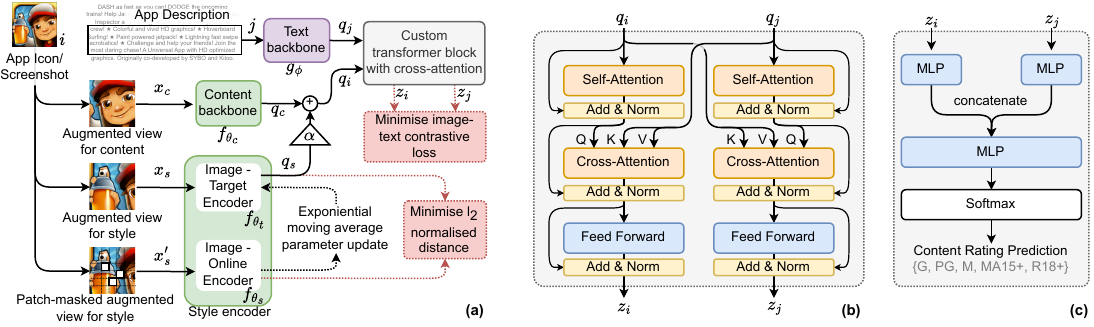}
    \caption{(a): Vision-language model architecture during the training stage. (b): Custom transformer block with cross-attention. (c) Pipeline for the downstream task of content rating classification.}
    \label{fig:model_architecture}
    % \vspace{-0.32cm}
\end{figure*}

% \vspace{-0.4cm}
\section{Methodology}
\label{Sec:methodology}

We propose a customised vision-language (VL) model that can be trained end-to-end with image and language data for learning joint representations directly via image patches and raw text tokens. Fig.~\ref{fig:model_architecture}(a) gives an overview of our model. It uses mobile app creatives such as app icons/screenshots and app descriptions as the paired inputs to generate joint embeddings that are useful for the downstream task of content rating prediction. We discuss our dataset in detail in Sec.~\ref{Sec:dataset}.

At a high level, we adapt two image encoder-based backbones to learn image content and style features separately. These two encoded features are merged together as image features. Then, we employ a text backbone to encode text features in the corresponding image-text pair belonging to an app. A cross-modal module then extracts relationships between image and textual features to produce the final image and textual embedding representations. We use pair-wise Sigmoid contrastive loss to learn the model parameters.

\subsection{Encoding Visual Information}
\label{sec:visual_info}
\begin{figure}[th]
    \centering
    \includegraphics[width=0.97\linewidth]{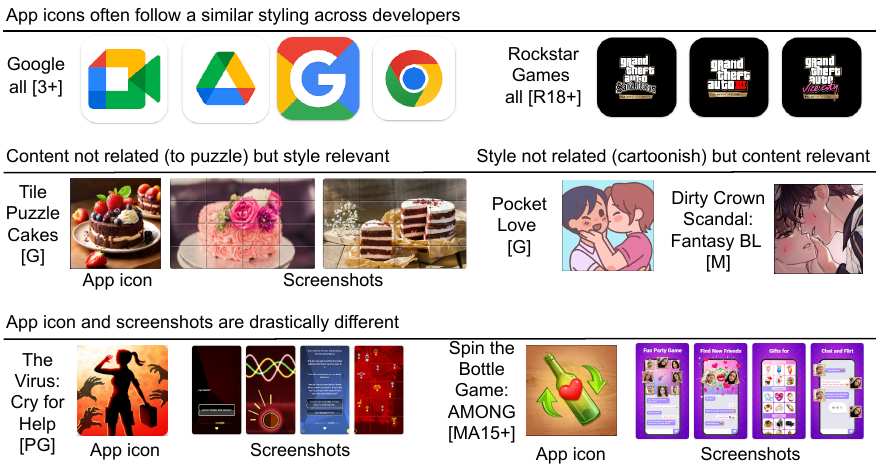}
    \caption{Disparity between the content and style of app icons and screenshots.}
    \label{fig:style_content_difference}
    % \vspace{-0.5cm}
\end{figure}

App icons and screenshots are the most prominent static-visual information the prospective app users first observe. In some cases, the style of an app icon alone is enough to distinguish many popular apps. As an example, app icons from Google would likely contain four colours: red, yellow, green and blue (\textbf{cf.} Fig.~\ref{fig:style_content_difference}). However, encoding such information 
is challenging as they do not contain features a generalised encoder such as CLIP was pre-trained on. To address this, we introduce trainable content and style encoder modules that work together to generate the final image embedding vector. Their ablation studies are further discussed in Sec.~\ref{Subsec: ablations}.

\subsubsection{Style Encoder}
\label{sec:style_enc}

Style information (e.g., texture, colour schemes, artistic choices) is crucial when predicting content rating as it adds context beyond the objects or elements present in images.
Kid-specific apps often use cartoonish characters, which are often associated with distinctive shapes and distinctive colour palettes and textures (e.g., bright colours, glitter texture, non-complex surface reflections), while dark tones (e.g. blood) or provocative lighting would be inappropriate for younger audiences, even if the content seems neutral~\cite{liu2016identifying}.

In Fig.~\ref{fig:style_content_difference}, we provide some example cases of apps with a significant disparity between the content and style of their creatives relative to the app category and content rating. The app icon and screenshots for \emph{The Puzzle Cakes} showcase cake related content and would have to rely on the style to associate it with a puzzle game. Conversely, \emph{Pocket Love} and \emph{Dirty Crown Scandal} employ a similar animated, cartoonish style but are aimed at distinctly different audiences based on their content ratings. Furthermore, there can be inter-app content and style disparities as in the examples of  \emph{The Virus} and \emph{Spin the Bottle Game}. In both cases, the colour themes of the app icons are very different from the screenshots.  

Therefore, we embedded a separate style encoder module. We use the CLIP image encoder as the base network for the style encoder, employing a masked representation learning task. This involves two identical networks, as illustrated in Fig.\ref{fig:model_architecture} (a). 
First, we uniformly sample an image $i$ from the dataset $D$ and generate an augmented image $x_s = t_s(i)$ by applying an image augmentation $t_s \sim T$. Next, we randomly mask three $3\times3$ patches to produce the masked image $x_s^{'}$.
The masked image is provided for one network called \emph{online}, while the unmasked image passes through another network called \emph{target}, a slow-moving average network. This allows the network to focus on the features that are invariant to masking. The target network $\theta_t$ uses an exponential moving average (EMA) of the online network $\theta_s$ to learn lower semantic features~\cite{assran2022masked, he2022masked}. More precisely, given a target decay rate $\tau \in [0, 1]$, after each training step, we update the target network weights using,
\begin{equation}
\label{eq:ema}
 \theta_t \leftarrow \tau \theta_t + (1 - \tau) \theta_s.
\end{equation}
The EMA introduces stability by averaging the network’s weights over time, smoothing out rapid updates that occur during the training process. This slows down the learning of fast-changing, higher-level features and enables the capturing of lower-level feature. From the masked image $x_s^{'}$, the online network outputs a representation $q_s^{'} = f_{\theta_s}(x_s^{'})$. The target network outputs $q_s = f_{\theta_t}(x_s)$ from the augmented view $x_s$. Finally we define the mean squared error between the embeddings $q_s$ and $q_s^{'}$ 
\begin{equation}
\label{eq:l2}
 \mathcal{L}_{mse} = ||q_s - q_s^{'}||_2^2.
\end{equation} 
and is added to the final loss, which we discuss later in Sec.~\ref{subsec: loss fn}. Furthermore, the generated style embeddings from the target network are scaled down using the hyperparameter $\alpha$ and added to the embeddings generated by the content backbone. 

\subsubsection{Content Encoder}
As shown in Fig.~\ref{fig:model_architecture}(a), our image content backbone, $f_{\theta_c}$, is built on the CLIP image encoder and remains active throughout the training process. This branch operates on augmented images of 224x224 resolution, which could be derived from an app icon or screenshots. We follow the augmentation settings defined in CLIP to generate different views of the image. The content backbone outputs visual content features, $q_c$ and are combined with visual style features, $q_s$ given by the style encoder. We represent this combination using the equation $q_i=q_c+\alpha. q_s$ where $\alpha$ is empirically selected as $0.1$. 

\subsection{Encoding Textual Information}
The app description provides an overview of the app's functionalities, features, and content to the prospective audience. Often, the text is summarised as app users are reluctant to read lengthy texts (capped at 4,000 characters; the average length of an app description in the top 20,000 apps of Google Play Store is 2,169 words), and Google Play mandated it to be general audience friendly. We perform randomised text chunking with four or more consecutive sentences randomly extracted from the long app description to be paired with respective app visuals.
We used a 110 million parameter RoBERTa text transformer with maximum $256$ tokens marked as text backbone in Fig.~\ref{fig:model_architecture}(a) to encode this information while the model parameters are kept frozen during the training. 

\subsection{Image-Text Cross Attention}
\label{ssec:cross attention}

Typical image captioning datasets such as MS-COCO~\cite{lin2014microsoft} and Flickr30k~\cite{young2014image} have a strong correlation between the captions and the images.
In contrast, app icons/screenshots and descriptions can exhibit a larger semantic gap, especially in the context of the content rating prediction problem because: 1) the description may not perfectly reflect what is depicted in the app icon or screenshots, 2) these modalities may not always contain useful information related to the content/age rating, and 3) App images display complex variations within the same content rating class. Therefore, we employ a stack of cross attention layers~\cite{vaswani2017attention} to align visual and textual tokens to fill the correlation gap between image patches and words. As shown in Fig.~\ref{fig:model_architecture}(b), our custom cross-attention module initially has a self-attention layer followed by a cross-attention layer. This layer induces text information to the image features. The \emph{query} values are derived from the previous image content encoder layer, and the memory \emph{keys} and \emph{values} are obtained from the hidden layers of the text encoder and vise versa. Text-to-image cross attention allows every patch of the image to attend over all tokens in the input sequence. Conversely, in \emph{image-to-text cross attention}, the roles are reversed, allowing every token of the input sequence to attend over all patches in the content image.
Finally, we introduce an additional linear projection layer, which outputs the final visual and textual embeddings.

\vspace{-0.15cm}
\subsection{Loss Function}
\label{subsec: loss fn}

Given a paired image and text sample $(i, j)$ from dataset $D$, two transformations $t_c$ and $t_s$ are drawn from a distribution of image augmentation $T$ (cf.~\ref{sec:visual_info}), to produce two distinct views $x_c = t_c(i)$ and $x_s = t_s(i)$ of the image $i$. These views serve as inputs to the image content backbone $f_{\theta_c}$ and the image target encoder $f_{\theta_t}$ in the style encoder block, respectively.
The views $x_c$ and $x_s$ are first encoded by $f_{\theta_c}$ and $f_{\theta_t}$ into their representations $q_c = f_{\theta_c}(x_c)$ and $q_s = f_{\theta_t}(x_s)$, which are then linearly combined to get the representation $q_i$. The text $j$ is encoded by a text backbone, $g_{\phi}$ into their representation $q_j = g_{\phi}(j)$. Then, these representations, $q_j$ and $q_i$ are mapped by the custom cross attention modules onto the embeddings $z_i$ and $z_j$. The Sigmoid Contrastive Loss~\cite{zhai2023sigmoid} is computed at the embedding level on $z_i$ and $z_j$. 

As defined in Eq.~\ref{eq:sigcl}, we adopt a supervised Contrastive Loss, more specifically, Sigmoid Contrastive loss (SigCL) proposed by~\cite{zhai2023sigmoid} with content rating label, where we consider image-text pairs with the same rating as positive pairs. This enables it to distinguish between data points not just based on data similarity but also according to their categories. Additionally, the SigCL benefits over Unified Contrastive Loss (UniCL)~\cite{yang2022unified} in a multi-modal setting because, when $N$ image-text pairs from the same content rating category are presented in a batch, UniCL is bounded and the maximum softmax value per pair is limited to $1/N$.  
Meanwhile, SigCL varies between 0 and 1 for each positive pair. We defined the SigCL between $z_i$ and $z_j$ embeddings along the batch $B$ as, 

\begin{multline}
\label{eq:sigcl}
    \mathcal{L}_{SigCL} =-\frac{1}{|P|} \sum_{i,j\in P} log \frac{1}{1+ e^{(-\tau z_i.z_j + b)}}\\
    -\frac{1}{|B|} \sum_{i\in|B|}\sum_{j\in|B|\backslash \{P\}} log \frac{1}{1+ e^{(\tau z_i.z_j - b)}}
\end{multline}

where $y_i$ and $y_j$ are the labels for a given image and text pair and $P = \{k|k\in B, y_i=y_j\}$, which represents the image text pairs coming from the same content rating. The $b$ in Eq.~\ref{eq:sigcl} alleviates the heavy imbalance coming from the many negatives. We also employ Euclidean distance loss, $L_{mse}$ to learn low level information such as texture and colour in image data. Specifically, we take the augmented image $x_s$ and generate a masked version of it. The two views are encoded by the $target$ and $online$ networks described in Sec.~\ref{sec:style_enc} into representations $y_s$ and $y'_s$, and optimise them using  $L_{mse}$. The entire network is optimized by minimising the following loss function:%\vspace{-0.3cm}
\begin{equation}
\label{eq:finallos}
    \mathcal{L} = \mathcal{L}_{SigCL}+\lambda \mathcal{L}_{mse}
\end{equation}
where $\lambda$ is a positive constant trading off the importance of the first and second terms in the loss L.

\section{Experimental Setup}

\subsection{Dataset}
\label{Sec:dataset}
Our dataset is a snapshot of the Google Play Store, which includes metadata and creatives for 1.3 million apps. This dataset was collected using a Python crawler from January 2023 to November 2023. We deployed an extremely slow crawling rate during this data collection.
For this work, we filtered out and used only the games category, which is more popular among children and as such, the correct content rating matters significantly. During our crawl, the crawler's geo-location was set as Australia (AU) to be consistent in obtaining content rating values of G, PG, M, MA15+, or R18+.

We sorted the selected gaming apps by rank, i.e., sorting in the descending order of number of downloads, star rating count and final star rating number following similar previous work~\cite{seneviratne2015early,rajasegaran2019multi}, and the first 20k games were selected as training and validation sets (80:20 random split) while the next 10k games were selected as the test set. We specifically did not mix the former due to the assumption that more popular apps are well monitored within the community and well maintained by the developers such that the metadata and content ratings information are less noisy than in the rest of the order. Due to the scarcity of games in categories of MA15+ and R18+ within the top 30k, we expanded our search space for them and appended them into train, validation and test sets. For analysis purposes, we created another dataset by including apps with the `Teacher Approved' tag~\cite{teacherAApps}. We report the distribution of apps by content rating across various datasets in Tab.~\ref{tab:dataset_distribution}.

\begin{table}[ht]
    \centering
    \caption{Dataset split and class distribution}
    % \vspace{-0.3cm}
    \label{tab:dataset_distribution}
    \begin{tabular}{l cccc}
    \hline
        & Train & Valid. & Test & Teacher Approved \\
        \hline
        G  & 4,544& 1,139&2,650& 2140\\
        PG & 4,540& 1,130&2,649& 30\\
        M  & 4,530& 1,120&2,648& 2\\
        MA15+ & 2,131& 547&1,796&- \\
        R18+ & 255& 62&255&- \\
        \hline
    \end{tabular}
    % \vspace{-0.3cm}
\end{table}

\begin{figure*}[ht]
    \centering
    \includegraphics[width=\linewidth]{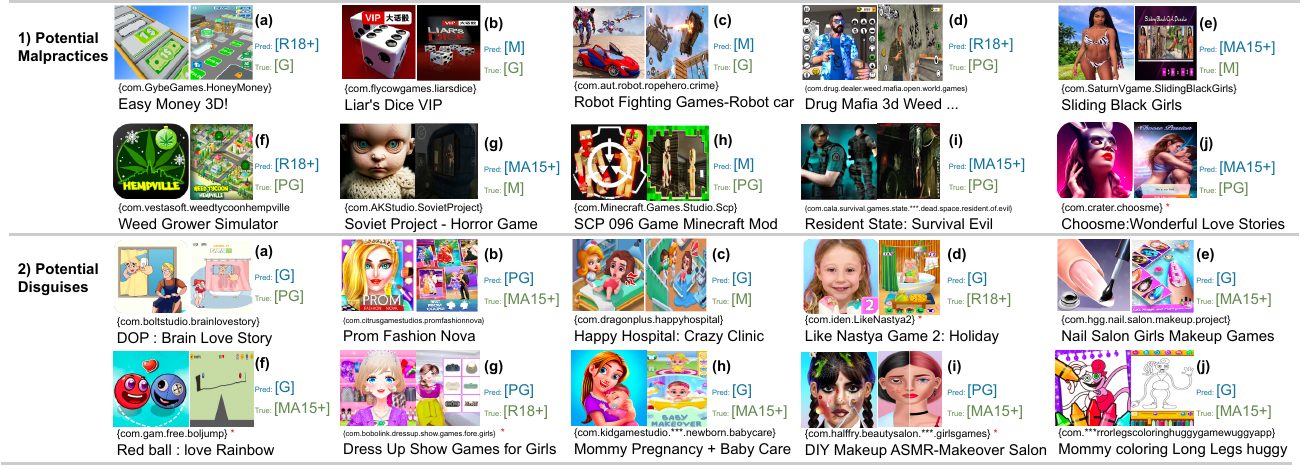}
    \caption{Examples belonging to 1) potential malpractices, and 2) potential disguises. For each app, the image on the left represents the app icon, and on the right is a screenshot. Red * represents app that removed from the Play store after the initial data crawl in 2023.}
    \label{fig:malpractices}
\end{figure*}

\subsection{Implementation Details}
\label{sec:ImplementationDetails}
We use the ViT-B/16 CLIP image encoder as our image backbone for both style and content branches and a frozen RoBERTa backbone in the text encoder branch. The model is pre-trained on eight NVIDIA V100 GPUs for 30 epochs with a minibatch size of 64. We use the Adam optimizer~\cite{kingma2014adam} with learning rate of $10^{-5}$, momentum of $0.9$, and weight decay of $0.02$. The learning rate follows a cosine decay schedule~\cite{loshchilov2016sgdr}, starting from 0 with $10$ warmup epochs and with a final value of $10^{-8}$. We perform a grid search to select the loss coefficients $\lambda$ in Eq.~\ref{eq:finallos} and set it to 5.

\subsection{Content Rating Predictions}
\label{subsec: content rating prediction}

We train a linear classification head to perform content rating predictions based on the previous outputs of image and text embeddings $z_i$ and $z_j$. That is, we propagate them via two separate MLP networks, concatenate the outputs, and then propagate again via another MLP network, followed by softmax classification to identify the prediction class. This stage is shown in Fig.~\ref{fig:model_architecture}(c). As we possess multiple images (app icon and screenshots) for a given app, we take the majority voting for the classification outputs for all of such image and text pairs. Note that we disable back-propagation in all the steps starting from $(i,j)$ up to obtaining $z_i,z_j$ during this classification stage. 

\subsubsection{Performance Metrics}

Due to the persistent class imbalance of our datasets, we report our model's performance in macro and weighted versions of \emph{precision}, \emph{recall} and \emph{F1 scores}. The overall \emph{accuracy} is calculated based on the elements of the principle diagonal of the confusion matrix. Predictions mapping to upper triangular or lower triangular portions of the confusion matrix are undesirable for app users and we later evaluate them in Sec.~\ref{subsec: potential malpractices} as \emph{potential malpractices} and in Sec.~\ref{subsec: potential disguises} as \emph{potential disguises}, respectively.

\section{Results}
\label{sec:results}

In this section, we first present the performance results of our model compared against several state-of-the-art baseline architectures representing image embeddings, text embeddings and image-text multi-modal embeddings.
Next, in the ablation study, we demonstrate the effect of different components of our approach by removing each component separately. Further, we evaluate the effect of using symmetric cross entropy loss (SCE)~\cite{clip}, which is used for CLIP pre-training, uniCL loss~\cite{yang2022unified} and SigCL loss~\cite{zhai2023sigmoid}.

\subsection{Performance Comparison with Baselines}

\begin{table}
    \centering
    \caption{ Performance Evaluation. Linear classification on top
of the frozen image and text representation and supervised baselines}
    \label{tab: performance comparison}
    % \vspace{-0.4cm}
    \begin{tabular}{llccccc}
    \hline
    &            & $P_m$ & $R_m$ & $P_w$ & $R_w$ & Acc \\ 
    
    \multicolumn{7}{l}{\emph{1) Image embeddings only}} \\ 
    \hline
    &   ResNet50    & 26.48 & \textbf{45.54} & 44.07 & 38.60 & 38.61\\
    &   ViT         & 28.33 & 35.25 & 45.18 & 46.06 & 46.07 \\
    &   BLIP        & 40.48 & 32.74 & 48.65 & 45.18 & 45.19 \\
    \cline{2-7}
    &   CLIP        & 56.88 & 42.57 & 51.24 & 50.02 & 50.03   \\
    &   CLIP-f.t.   & 56.64 & 42.39 & 51.01 & 49.80 & 49.79 \\
    &   Ours        & \textbf{61.18} & 45.14 & \textbf{53.10} & \textbf{51.14} & \textbf{51.15} \\
    \cline{2-7}
    % \vspace{-0.2cm}
    % &&&&&&\\
    \multicolumn{7}{l}{\emph{2) Text embeddings only}}\\
    \hline
    &   BERT        & 45.28 & \textbf{47.94}& 51.06 & 49.32 & 49.33 \\
    &   RoBERTa     & 52.35 & 46.86 & 49.82 & 49.83  & 49.84 \\
    &   BLIP        & 38.10 & 41.20 & 45.11 & 39.75 & 39.76 \\
    \cline{2-7}
    &   CLIP        & 51.08 & 42.62 & 47.44 & 47.65 & 47.46   \\
    &   CLIP-f.t.   & 49.25 & 41.63 & 46.69 & 47.08 & 50.18 \\
    &   Ours        & \textbf{60.58} & 43.31 & \textbf{52.54} & \textbf{50.45} & \textbf{50.46} \\
    \cline{2-7}
    % \vspace{-0.2cm}
    % &&&&&&\\
    
    \multicolumn{7}{l}{\emph{3) Image-Text embeddings only}}\\
    \hline
    &   ViT+RoBERTa & 33.33 & 33.87  & 48.45 & 50.78 & 50.78 \\
    &   BLIP        & 33.26 & 32.59 & 47.73 & 49.13 & 49.13 \\
    \cline{2-7}    
    &   CLIP        & 53.66 & 45.44 & 49.97 & 50.11 & 50.12   \\
    &   CLIP-f.t.   & 57.70 & 45.56 & 50.60 & 50.19 & 50.18 \\
    &   Ours        & \textbf{61.57} & \textbf{46.63} & \textbf{54.80} & \textbf{53.09} & \textbf{53.09} \\
    \hline
    
    \end{tabular} 
\end{table}

We evaluate the performance of our method against pre-trained and fine-tuned CLIP models in three settings as shown in Tab.~\ref{tab: performance comparison}; 1) using only CLIP image embeddings ($z_i$), 2) using only CLIP text embeddings ($z_j$), and 3) using both CLIP image and text embeddings ($z_i$ and $z_j$). For 1) and 2), we modify the experimental setup explained in Sec.~\ref{subsec: content rating prediction} such that either $z_i$ or $z_j$ would only propagate via a single MLP network before obtaining the softmax scores. In 1) and 3), we take the majority voting output considering all images that come with an app, but not in 2). This is because one app has only one text description.

\begin{figure}[h]   
    \centering
    \includegraphics[width=0.97\linewidth]
    {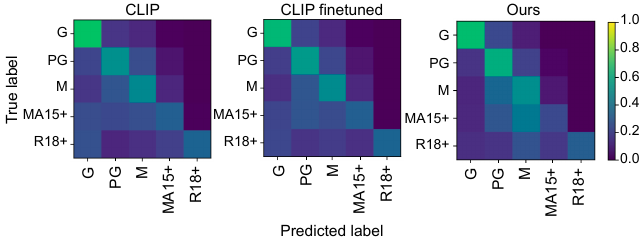}
    \caption{Confusion matrix comparing our method against baselines - using image-text embeddings.}
    \label{fig:confution_matrix}
\end{figure}

For the pre-trained CLIP model baselines, we initialise the original parameters for the model and freeze them before training the content rating classifier. For fine-tuning CLIP, we use our training dataset consisting of app icons, screenshots and the respective app descriptions to fine-tune the already pre-trained CLIP model for an additional 13 epochs using a batch size of 384. Next, we freeze the fine-tuned model parameters to train the content rating classifier.

As detailed in Tab.~\ref{tab: performance comparison}, our model achieves an accuracy of $\sim$53.1\% in the image-text embedding classifier setting and outperforms the pre-trained CLIP by a relative percentage increase of $\sim$$5.9\%$ and the fine-tuned CLIP by a relative percentage increase of $\sim$$5.8\%$. We also report other metrics, such as precision and recall in macro ($m$) and weighted ($w$) settings for further comprehension. 
Despite a relative decrease in accuracy by 3.7\% for the image-only classifier and 5.0\% for the text-only classifier, our methods outperform CLIP and CLIP-ft across all evaluated metrics. We attribute this improved performance to the integration of cross-attention mechanisms and a style encoder, which enable our model to learn richer visual and textual features compared to models leveraging two separate encoders for each modality.

We further portray the confusion metrics in Fig.~\ref{fig:confution_matrix}, comparing our method against baselines of CLIP and CLIP-ft, where our predictions are less scattered off-diagonal in the upper and lower triangles. 
Based on these results, we can conclude that our method outperforms both CLIP and CLIP-ft among all three settings depicted in Tab.~\ref{tab: performance comparison} and the best results are obtained when both image and text embeddings are considered.

For completeness, we also consider standard image and text baselines that have been supervised fine-tuned on our training dataset. These models are trained end-to-end using app creatives (app icons/screenshots and app descriptions ) to predict content rating labels directly. 
Specifically, we present image-only classification results using ResNet50~\cite{he2016deep}, ViT~\cite{dosovitskiy2020image}, and BLIP~\cite{li2022blip} image encoders as the backbone of the classifiers. For text classification results, we use BERT, RoBERTa and BLIP text encoder-based classifiers. For image-text classification results, we use a concatenation of ViT+RoBERTa encoder and BLIP text and image concatenated encoder-based classifiers.

ResNet50 and ViT both perform poorly compared to our method on average by 20.2\%. Comparatively, BERT and RoBERTa only demonstrate an average accuracy drop of $\sim$6.6\%. This disparity in performance can be attributed to several factors. App descriptions sometimes provide explicit, detailed information about the content, usage, and target audience, directly reflecting the attributes relevant to content rating. 

ViT and RoBERTa concatenated classifier has an accuracy of 50.78\% which is still 4.35\% under performing than our method. Additionally, BLIP image or text or image and text encoder classifiers are also under performing than our method in average by 15.81\%. These results again establish that, despite training an end-to-end model to perform content rating classification, it is not effective compared to our method in producing more meaningful embeddings that subsequently produce better content rating classifications.

\subsection{Ablation Studies}
\label{Subsec: ablations}

\noindent{\textbf{Ablations with respect to loss functions:}
We alter our contrastive loss function in Eq.~\ref{eq:sigcl} in several ways to experiment with different loss functions in SSL, such as SCE loss, UniCL, and SigCL. We compare linear classification results trained on the combination of image and text representations. The results of Tab.~\ref{tab: loss_comparisson} show that our method trained on SigCL outperforms SCE and UniCL methods by 9.3\% and 3.0\%, respectively.} \\

\begin{table}
    \centering
    \caption{Performance with different losses}
    \vspace{-0.3cm}
    \label{tab: loss_comparisson}
    \begin{tabular}{llccc}
        \hline
        & Metric &SCE & UniCL &Ours\\
        \hline
        Macro & Precision & 56.36 & 60.29 & \textbf{61.57}\\
         & Recall &38.52& 43.45 & \textbf{46.63}\\
         & F1 Score &38.17 & 45.23 & \textbf{49.25}\\
         \hline
        Weighted & Precision &49.74 & 52.63 & \textbf{54.80}\\
         & Recall &48.59 & 51.51 &\textbf{53.09}\\
         & F1 Score & 46.77& 50.50 & \textbf{52.31}\\
        \cline{2-5}
          & Accuracy & 48.59 & 51.52 &\textbf{53.09}\\
        \hline      
    \end{tabular}   
    \vspace{-0.4cm}
\end{table}

\noindent{\textbf{Ablation of model components:} To observe the effect of the style encoder, we evaluate our model with only the content-encoder as presented in Tab.~\ref{tab: deactivation_test}. The macro average precision observes a gain of 1.67\% without the style encoder (i.e., less likely to make false predictions but recalls less: -6.58\%), yet, all the other metrics indicate better performance with our method (macro average F1 score: +7.43\%, accuracy: +4.65\%). Next, we augmented our methodology without cross-attention and replaced it with a self-attention block. Our method achieves better performance in all the metrics compared to this setting. Removing cross-attention disproportionately affects classes, as indicated by a larger drop in the weighted F1 score compared to the macro F1 score. This suggests that cross-attention is crucial for maintaining performance in majority classes like G, PG, and M, helping the model effectively distinguish between these content ratings. 
Overall, ablation study results show that style backbone added with text-image cross-attention contributes to the increased performance.}
         
\begin{table}
    \centering
    \caption{Effect of incorporating style encoder and cross attention}
    \label{tab: deactivation_test}
    \vspace{-0.3cm}
    \begin{tabular}{llccc }
    \hline

         & Metric &w/o style & w/o cross &Ours\\
         & &encoder& attention &\\
        \hline
        Macro & Precision & \textbf{62.60} & 58.81 & 61.57\\
         & Recall &43.56 & 46.43& \textbf{46.63}\\
         & F1 Score &45.84 & 49.12& \textbf{49.25}\\
         \hline
        Weighted & Precision &\textbf{54.91}& 50.99 & 54.80\\
         & Recall &50.72 &50.76 &\textbf{53.09}\\
         & F1 Score & 49.86&49.80 &\textbf{52.31}\\
        \cline{2-5}
          & Accuracy & 50.73&50.76  &\textbf{53.09}\\
        \hline      
    \end{tabular}  
    \vspace{-0.4cm}
\end{table}

\section{Result Analysis}

In this section, we analyse the results and predictions of our method from the perspective of the mobile app ecosystem. First, we observe how deviated the text and visual data are compared to natural language and text. Next, we present interesting findings among the lower and upper triangular parts of the confusion matrix (i.e., apps having a lower or higher content rating than our predictions).

\subsection{Image-text Cross Attention}
\begin{figure}[ht]  
    \centering
    \includegraphics[width=0.97\linewidth]{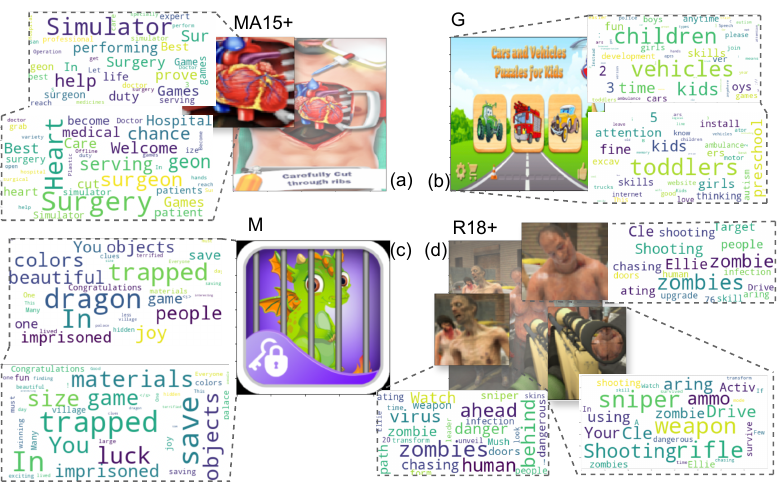}
    \caption{Visualisation of image patches attending to text tokens in the custom cross-attention layer.}
    \label{fig:ca_visualized}
\end{figure}

In Sec.~\ref{ssec:cross attention}, we discussed how our image-text cross attention (CA) design allows every patch of the content image to attend over all tokens in the text input sequence. Also, for each image patch, there are 12 attention heads running in parallel. Therefore, to qualitatively measure how the CA happens, we select attention-heads of the first layer as lower layers are often associated with broader attention~\cite{vig_2019_multiscale}. Then, for each image patch, we select the tokens with the highest numerical attention values after excluding some stop-word and punctuation mark-related tokens. In Fig.~\ref{fig:ca_visualized}, we visualise the highly attended words in a given input image portion for some example images. An image portion is a collection of consecutive patches which we select as a region of interest; for example, the patches outlining the gun in Fig.~\ref{fig:ca_visualized} (d) or the heart in Fig.~\ref{fig:ca_visualized} (a). The results show that mobile app ecosystem-related tokens such as `simulator', `game', `upgrade', and `developers' are now attended by the image patches, even though those words cannot be identified by observing the image in a general context. Furthermore, the tokens representing target audiences such as `kids', `children' and `girls' are now attended. This further demonstrates how our model has been able to reduce the gap between visual data and textual data in the mobile app domain.

\subsection{Predictions in the Wild}

When our classifier is applied to the test set containing 10,000 gaming apps, two interesting cases emerge, i.e., apps with a higher or lower labelled rating than our model's predictions. These two scenarios lead to potential  \emph{``malpractices"} and \emph{``disguises"} that are important from a content safety point of view. 

\subsubsection{Potential malpractices}
\label{subsec: potential malpractices}

When our method predicts a class label that is higher than the developer-defined label, we characterise such examples as possible malpractices (i.e., having a lower content rating than what the app is supposed to have). Occurrences belonging to this category could be identified along the horizontal axis of the confusion matrix, on the right-hand side to the diagonal. This is more observed in the G category as any higher prediction, such as PG, M, MA15+, and R18+, raises a concern. As we go higher in prediction classes, the possibilities for malpractices decrease, and therefore, the R18+ category is not susceptible to malpractices. 

We manually evaluated 350 apps with predicted labels that were two classes or more higher than the true label (e.g., prediction [M, MA15+ or R18+] when the true label is [G]) and identified 62 (17.7\%) of them as potential malpractices (i.e., possible content rating violations). \textit{Also, we highlight that at the time of writing, 14 of them were removed from Play Store, and 20 of them were increased to a higher rating class compared to the time we crawled the dataset. While we can't be exactly sure why these 14 apps were removed from  Google Play Store, previous work has reported that Google take down apps violating their content policies~\cite{seneviratne2015early}.} 

We highlight ten exemplary instances that are potentially linked to malpractices in Fig.~\ref{fig:malpractices} (1). Examples (a) and (b) both represent gambling-related games rated [G] that would at least require a rating of [PG]. Example (c) depicts shooting and gun usage, again not suitable for a general audience. (e) and (j) contain images more suited for an [MA15+] audience, and (d) and (f) entertain visual and textual cues strongly related to illegal substances that require a rating higher than [PG]. (g) and (h) contain images with horror themes, blood and intense cartoon or fantasy violence which are more suited for [MA15+] audience. These examples suggest that our model can flag potential content malpractices in the app ecosystem.

\subsubsection{Potential disguises}
\label{subsec: potential disguises}

Apps belonging to higher content ratings are likely not to conduct malpractices, but alarmingly, they could be disguised as a lower-rated app and, as a consequence, could attract an unsuitable audience to the app.  
As an example, an R18+ app consisting of cartoonish images and not-so-alarming textual data could attract underage audiences due to their natural tendency for curiosity and their interest in cartoons. From the developer's perspective, they comply with the content policies and applicable laws. However, in such cases, we argue that at least the textual description must contain information such that someone (e.g., a parent or a guardian) who misses the content rating label should be able to figure out the app's purpose and functionality independently. We define this category as possible disguises. R18+ is more vulnerable to disguises that can be identified horizontally left to the diagonal of the confusion matrix. Being the lowest content rating, category [G] is not susceptible to disguises. We checked 131 samples that were predicted [G, PG, M] when the true class was [R18+], and 203 samples predicted [G] when the true class was [MA15+] and observed $\sim$9.3\% of them to be possible disguises.

We demonstrate some of such examples in Fig.~\ref{fig:malpractices}(2). App represented by (a) is an app rated for [PG] and our method predicts even lower as [G]. Observing the images, it is evident their content is sexual in nature despite the cartoonish theme. In our test dataset, the average rating for an app with such sexualised imagery is [MA15+]. 
Note that in this category, our model is likely to predict a lower rating as the images or text is not suggestive of requiring high ratings. Due to such examples being rare, our model does not know how to predict them correctly, yet we still can automatically identify them based on our off-to-left-diagonal results, as explained before. Apps (b, e, g) and (i), though rated for [MA15+] and higher based on a storyline related to a `fashion and makeup', is likely to attract a younger audience due to the visual appearance similar to the majority of [PG] rated \emph{Dress up} games. A similar interpretation can be given to [M] rated examples (c, h), which are too cartoonish yet contain images related to mature audiences, such as pregnancy. Furthermore, (f) and (j) are rated as [MA15+], which contains appealing games for young audience. Despite measures such as \emph{Not designed for children} tagging is available in Google Play~\cite{notdisgisechil.} to safeguard users, it doesn't appear to be used by many developers. 

\subsubsection{Unverifiable apps} 
\label{subsec: unverifiable_apps}

\begin{figure}[ht]  
    \centering
    \includegraphics[width=0.97\linewidth]{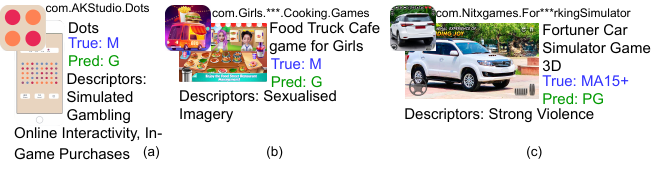}
    \caption{Examples of unverifiable apps with developer defined content rating descriptors.}
    \label{fig:unverifiable_apps}
\end{figure}
We further highlight discrepancies in content rating declarations, as illustrated in Fig.~\ref{fig:unverifiable_apps}. These noisy instances represent cases where we could not manually verify the developer-assigned ratings based on the app’s available metadata, visuals, or descriptions. The app in (a) is rated as [M] with descriptors indicating simulated gambling, online interactivity, and in-game purchases. However, compared to similar games in the dataset, it does not display any visual cues or textual descriptions to justify such a rating. Therefore, the absence of gambling graphics or explicit mention of gambling mechanics justifies the predicted rating of [G].
Similarly, app (b) and (c) fail to reflect sensitive content such as sexualised imagery and strong violence in app creatives. Hence, in all these cases, our model predicted a lower rating than the original rating. While it may not be explicitly illegal to omit sensitive descriptors from app screenshots or descriptions, missing descriptors in visuals hinder the user's ability to make informed decisions, leading to unwanted downloads or unexpected experiences and children and vulnerable users may unintentionally be exposed to harmful or age-inappropriate content.

\subsubsection{Teacher Approved (TA) Apps}

\begin{figure}[ht]  
    \centering
    \includegraphics[width=0.97\linewidth]{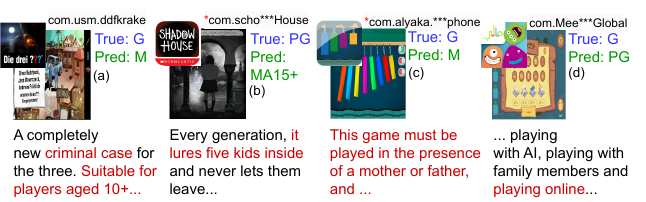}
    \caption{Examples of teacher-approved apps with incorrect content ratings - A section of the app description is quoted and * indicates this app is no longer available.}
    \label{fig:teacher_apps}
\end{figure}

Google Play has deployed the `teacher approved' apps based on consultations with experts to determine the suitability for kids, and especially the age appropriateness~\cite{teacherAApps}. Hence, these apps are likely to be more content appropriate as per the rating labelling. However, after analysing 2,172 TA apps, we found that our model predictions deviate from the declared content rating classifications, with 10.4\% of them being flagged for potential malpractices. Among them, 92\% of the instances were flagged as requiring [PG] despite being declared as [G]. Further evaluating their continuity, within a time span of nine months, we observed that 34.5\% of apps that were classified as potential malpractices have been removed from the Play Store. On the contrary, only 27.4\% of correctly predicted apps were removed, which is lower compared to the deletion rate of apps identified with malpractices.
We further discuss why app removal rate can be a proxy measure of content policy violations in the next subsection.

As depicted in Fig.~\ref{fig:teacher_apps} we manually verified a portion of apps flagged before as malpractices based on the available metadata and identified that nine apps are likely to be not suitable for children despite being tagged as TA. 
The presence of violence (Fig.~\ref{fig:teacher_apps}a), horror themes (Fig.~\ref{fig:teacher_apps}b) or online multiplayer interactions (Fig.~\ref{fig:teacher_apps}d) were the main reasons we identified behind these content rating discrepancies. 
The example in Fig.~\ref{fig:teacher_apps} c highlights a contradiction in the app's description, which mandates parental presence, despite the app being labeled as [G] in the Play Store.
\textit{Overall, the presence of these practices among `teacher approved' apps is alarming. It shows that even manually verified apps are not immune to content rating malpractices, and further rigour is required in app vetting.}

\begin{figure}[ht]    
    \centering
    \includegraphics[width=0.97\linewidth]{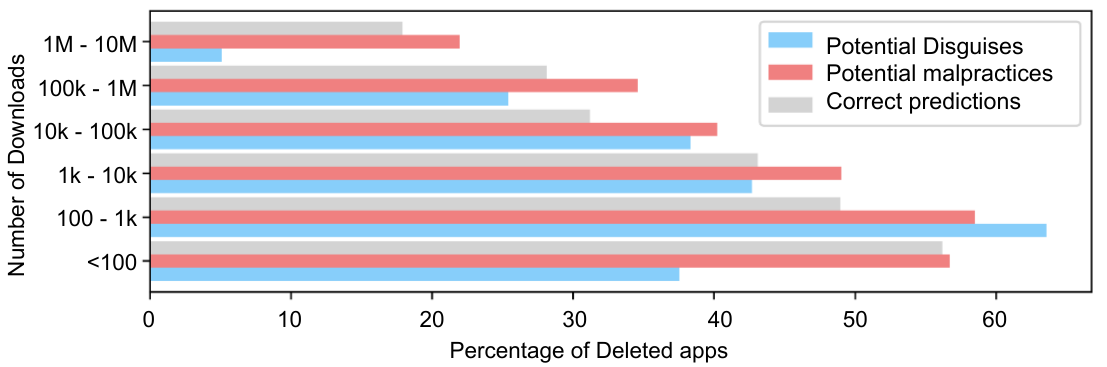}
    \caption{App deletion rates w.r.t number of downloads.}
    \label{fig:app_deletion}
   % \vspace{-0.3cm}
\end{figure}

\subsubsection{App Deletion Rate}

An app can be discontinued in Google Play for two reasons: the developer could discontinue the app~\cite{denipitiyage2024detecting}, or Google could remove the app for violating its policies~\cite{2023gplypolicyviolation}. As a result, an app being removed from Google Play can be used as an indication of a possible violation of Google Play policies.

To this end, we used 15,985 apps that are gathered from the test set (10,000 apps: c.f. Sec.~\ref{Sec:dataset}) added with 5,985 apps with lower downloads (download count $<$ 100,000 - to account for a better distribution as our test set consists of top apps only). Next, we attempted to re-crawl these apps to check whether they were still there in Google Play. Overall, we found 45.7\% of apps identified as having malpractices, 39.1\% of apps that predicted to be disguises were removed within the time span of nine months. In comparison only  29.1\% of apps that correctly predicted were removed.

In Fig.~\ref{fig:app_deletion}, we show the percentage of apps that we found as deleted according to the download numbers and the predictions of our classifier. At all download ranges apart from `$<$ 100', we notice that apps we classified as potential malpractices have a higher deletion rate than apps we classified as correct. Similar values of `$<$~100' category can be explained by less attention and consequently fewer complaints received on those apps for Google to action.

On the other hand, apps classified as disguises are likely not to be removed as malpractices, as they are unlikely to be noticed or complained about by an average audience. Notably, apps flagged as disguises with more than 1M downloads are far less likely to be removed as the number of apps with a higher content rating label (e.g., MA15+, R18+) are not frequent among the apps.

\section{Concluding Remarks}

\label{Conclusion}

In this paper, we proposed a vision-language approach to detect content rating violations in Google Play Store. Our model includes multiple trained encoders capturing features related to app creative styles, content, text descriptions and their relationships using a cross-attention module. We trained our model using a large dataset from Google Play Store focusing on gaming apps. Our method outperformed the state-of-the-art CLIP model, even when fine-tuned on the same dataset. We achieved 5.9\% and 5.8\% relative improvements in accuracy compared to CLIP and fine-tuned CLIP, respectively. Even though our method doesn't achieve perfect accuracy, apps that deviate from the predicted rating (i.e., potential malpractices or disguises) can serve as a shortlist for e-safety regulators and app market operators, thereby reducing manual effort. By leveraging static information such as images and text, we can quickly identify apps for further inspection. Beyond these two categories, we also identified unverifiable apps that have been assigned higher content ratings, yet even human reviewers could not justify these ratings based on the app creatives and descriptions alone. While there is no legal requirement for an app’s creatives to explicitly reflect its content rating, we emphasize the importance of ensuring alignment between content descriptors and app creatives/descriptions. This transparency will help users to make informed decisions.

We applied our model in the wild and found that our model can detect content rating malpractices in practice. We could identify 71 ($\sim$17\% of what we verified) of such examples.
Some notable examples include gambling apps such as \emph{Liar's Dice VIP} being categorised as G and \emph{Drug Mafia 3d Weed} being categorised as PG. In addition to that, within our test set, 16.86\% of the apps we identified as violating content policies are no longer available in Google Play Store due to potential banning, further justifying the effectiveness of our method. As an artefact, we also found another interesting behaviour related to app content ratings in the Google Play Store characterised as \emph{potentially disguises}. These apps have a correct content rating. However, their look and feel appear to target a general audience. For example, an app with a cartoonish theme but mature content may inadvertently attract children, for whom the content could be disturbing. We identified 32 such instances. Finally, we conducted an extended evaluation on 2,172 `teacher approved' apps and identified nine apps with possible content rating malpractices.

One limitation of our work is the reliance on top apps having reliable content ratings and representative app creatives. While several comparable works have used similar ideas in domains such as spam app detection~\cite{seneviratne2015early}, counterfeit detection~\cite{karunanayake2020multi}, this approach may introduce noise into the CLIP fine-tuning process. One way to mitigate this limitation is through human annotation, though this can be costly. Another approach is to match apps between the Google Play Store and Apple App Store using a method such as ~\cite{steinbock2024comparing}, leveraging Apple's content ratings, which typically undergo manual verification. However, Google and Apple use different content rating scales, which may introduce inconsistencies.

Additionally, incorporating other app metadata—such as user comments, data safety declarations, and dynamic app behaviors—could enhance the framework’s robustness against unverifiable apps. However, this approach is more resource- and time-intensive than analyzing text descriptions and app creatives. As a result, it could serve as a secondary classifier after identifying potential content rating violations at scale using our proposed method.

\section{Acknowledgment}
This research was supported by the Australian Government through the Australian Research Council’s Discovery Projects funding scheme (Project ID DP220102520).

% \newpage
\bibliographystyle{IEEEtran}
\bibliography{biblio}

\begin{thebibliography}{10}
\providecommand{\url}[1]{#1}
\csname url@samestyle\endcsname
\providecommand{\newblock}{\relax}
\providecommand{\bibinfo}[2]{#2}
\providecommand{\BIBentrySTDinterwordspacing}{\spaceskip=0pt\relax}
\providecommand{\BIBentryALTinterwordstretchfactor}{4}
\providecommand{\BIBentryALTinterwordspacing}{\spaceskip=\fontdimen2\font plus
\BIBentryALTinterwordstretchfactor\fontdimen3\font minus \fontdimen4\font\relax}
\providecommand{\BIBforeignlanguage}[2]{{%
\expandafter\ifx\csname l@#1\endcsname\relax
\typeout{** WARNING: IEEEtran.bst: No hyphenation pattern has been}%
\typeout{** loaded for the language `#1'. Using the pattern for}%
\typeout{** the default language instead.}%
\else
\language=\csname l@#1\endcsname
\fi
#2}}
\providecommand{\BIBdecl}{\relax}
\BIBdecl

\bibitem{ashtuener2024}
\BIBentryALTinterwordspacing
A.~Turner, ``Number of smartphone users worldwide (billions),'' 2024. [Online]. Available: \url{https://www.bankmycell.com/blog/how-many-phones-are-in-the-world\#part-1}
\BIBentrySTDinterwordspacing

\bibitem{liu2016identifying}
M.~Liu, H.~Wang, Y.~Guo, and J.~Hong, ``Identifying and analyzing the privacy of apps for kids,'' in \emph{Proceedings of the 17th International Workshop on Mobile Computing Systems and Applications}, 2016, pp. 105--110.

\bibitem{2019smartphone}
\BIBentryALTinterwordspacing
``The common sense census: Media use by tweens and teens,,'' 2019. [Online]. Available: \url{https://www.commonsensemedia.org/research/the-common-sense-census-media-use-by-tweens-and-teens-2019}
\BIBentrySTDinterwordspacing

\bibitem{teacherAApps}
``Build teacher approved apps,'' \url{https://play.google.com/console/about/programs/teacherapproved/}, accessed: 2024-08-07.

\bibitem{luo2020automatic}
Q.~Luo, J.~Liu, J.~Wang, Y.~Tan, Y.~Cao, and N.~Kato, ``Automatic content inspection and forensics for children {Android} apps,'' \emph{IEEE Internet of Things Journal}, vol.~7, no.~8, pp. 7123--7134, 2020.

\bibitem{anualdatareport2023}
\BIBentryALTinterwordspacing
Qustodio, ``Qustodio releases 2023 annual report, born connected: The rise of the ai generation,'' 2023. [Online]. Available: \url{https://static.qustodio.com/public-site/uploads/2024/01/19122535/ADR_2023-24_EN.pdf}
\BIBentrySTDinterwordspacing

\bibitem{sun2023not}
R.~Sun, M.~Xue, G.~Tyson, S.~Wang, S.~Camtepe, and S.~Nepal, ``Not seen, not heard in the digital world! measuring privacy practices in children’s apps,'' in \emph{Proceedings of the ACM Web Conference 2023}, 2023, pp. 2166--2177.

\bibitem{content_ratin_q_geographical}
``Developer content policy,'' \url{https://support.google.com/googleplay/android-developer/answer/9859655?sjid=12738238277494853083-AP&visit_id=638689148734509737-3391379066&rd=1#questionnaire}, accessed: 2024-12-05.

\bibitem{avada}
\BIBentryALTinterwordspacing
{Sam Nguyen}, ``{Android}’s latest statistics 2024: How many people have {Android}s?'' September 06, 2023. [Online]. Available: \url{https://avada.io/articles/how-many-people-have-androids/}
\BIBentrySTDinterwordspacing

\bibitem{app_review_apple}
``Developer content policy,'' \url{https://developer.apple.com/app-store/review/guidelines/}, accessed: 2024-12-05.

\bibitem{federal2012mobile}
F.~T. Commission, ``Mobile apps for kids: Current privacy disclosures are disappointing,'' \emph{Washington, DC. RetrievedAugust}, vol.~21, p. 2022, 2012.

\bibitem{nguyen2022freely}
T.~T. Nguyen, M.~Backes, and B.~Stock, ``Freely given consent? studying consent notice of third-party tracking and its violations of gdpr in {Android} apps,'' in \emph{Proceedings of the 2022 ACM SIGSAC Conference on Computer and Communications Security}, 2022, pp. 2369--2383.

\bibitem{nguyen2021measuring}
T.~T. Nguyen, D.~C. Nguyen, M.~Schilling, G.~Wang, and M.~Backes, ``Measuring user perception for detecting unexpected access to sensitive resource in mobile apps,'' in \emph{Proceedings of the 2021 ACM Asia Conference on Computer and Communications Security}, 2021, pp. 578--592.

\bibitem{hu2015protectingcikm}
B.~Hu, B.~Liu, N.~Z. Gong, D.~Kong, and H.~Jin, ``Protecting your children from inappropriate content in mobile apps: An automatic maturity rating framework,'' in \emph{Proceedings of the 24th ACM International on Conference on Information and Knowledge Management}, 2015, pp. 1111--1120.

\bibitem{liccardi2014can}
I.~Liccardi, M.~Bulger, H.~Abelson, D.~J. Weitzner, and W.~Mackay, ``Can apps play by the coppa rules?'' in \emph{2014 Twelfth Annual International Conference on Privacy, Security and Trust}.\hskip 1em plus 0.5em minus 0.4em\relax IEEE, 2014, pp. 1--9.

\bibitem{chen2013isthisapp}
Y.~Chen, H.~Xu, Y.~Zhou, and S.~Zhu, ``Is this app safe for children? a comparison study of maturity ratings on {{Android}} and {iOS} applications,'' in \emph{Proceedings of the 22nd international conference on World Wide Web}, 2013, pp. 201--212.

\bibitem{zhou2022automatic}
C.~Zhou, X.~Zhan, L.~Li, and Y.~Liu, ``Automatic maturity rating for {Android} apps,'' in \emph{Proceedings of the 13th Asia-Pacific Symposium on Internetware}, 2022, pp. 16--27.

\bibitem{chen2020simple}
T.~Chen, S.~Kornblith, M.~Norouzi, and G.~Hinton, ``A simple framework for contrastive learning of visual representations,'' in \emph{International conference on machine learning}.\hskip 1em plus 0.5em minus 0.4em\relax PMLR, 2020, pp. 1597--1607.

\bibitem{zbontar2021barlow}
J.~Zbontar, L.~Jing, I.~Misra, Y.~LeCun, and S.~Deny, ``Barlow twins: Self-supervised learning via redundancy reduction,'' in \emph{International conference on machine learning}.\hskip 1em plus 0.5em minus 0.4em\relax PMLR, 2021, pp. 12\,310--12\,320.

\bibitem{clip}
A.~Radford, J.~W. Kim, C.~Hallacy, A.~Ramesh, G.~Goh, S.~Agarwal, G.~Sastry, A.~Askell, P.~Mishkin, J.~Clark \emph{et~al.}, ``Learning transferable visual models from natural language supervision,'' in \emph{International conference on machine learning}.\hskip 1em plus 0.5em minus 0.4em\relax PMLR, 2021, pp. 8748--8763.

\bibitem{align}
C.~Jia, Y.~Yang, Y.~Xia, Y.-T. Chen, Z.~Parekh, H.~Pham, Q.~Le, Y.-H. Sung, Z.~Li, and T.~Duerig, ``Scaling up visual and vision-language representation learning with noisy text supervision,'' in \emph{International conference on machine learning}.\hskip 1em plus 0.5em minus 0.4em\relax PMLR, 2021, pp. 4904--4916.

\bibitem{oord2018representation}
A.~v.~d. Oord, Y.~Li, and O.~Vinyals, ``Representation learning with contrastive predictive coding,'' \emph{arXiv preprint arXiv:1807.03748}, 2018.

\bibitem{LiT}
X.~Zhai, X.~Wang, B.~Mustafa, A.~Steiner, D.~Keysers, A.~Kolesnikov, and L.~Beyer, ``Lit: Zero-shot transfer with locked-image text tuning,'' in \emph{Proceedings of the IEEE/CVF Conference on Computer Vision and Pattern Recognition}, 2022, pp. 18\,123--18\,133.

\bibitem{fang2021clip2video}
H.~Fang, P.~Xiong, L.~Xu, and Y.~Chen, ``Clip2video: Mastering video-text retrieval via image clip,'' \emph{arXiv preprint arXiv:2106.11097}, 2021.

\bibitem{portillo2021straightforward}
J.~A. Portillo-Quintero, J.~C. Ortiz-Bayliss, and H.~Terashima-Mar{\'\i}n, ``A straightforward framework for video retrieval using clip,'' in \emph{Mexican Conference on Pattern Recognition}.\hskip 1em plus 0.5em minus 0.4em\relax Springer, 2021, pp. 3--12.

\bibitem{nichol2021glide}
A.~Nichol, P.~Dhariwal, A.~Ramesh, P.~Shyam, P.~Mishkin, B.~McGrew, I.~Sutskever, and M.~Chen, ``Glide: Towards photorealistic image generation and editing with text-guided diffusion models,'' \emph{arXiv preprint arXiv:2112.10741}, 2021.

\bibitem{massiceti2023explaining}
D.~Massiceti, C.~Longden, A.~Slowik, S.~Wills, M.~Grayson, and C.~Morrison, ``Explaining clip's performance disparities on data from blind/low vision users,'' \emph{arXiv preprint arXiv:2311.17315}, 2023.

\bibitem{agarwal2021evaluating}
S.~Agarwal, G.~Krueger, J.~Clark, A.~Radford, J.~W. Kim, and M.~Brundage, ``Evaluating clip: towards characterization of broader capabilities and downstream implications,'' \emph{arXiv preprint arXiv:2108.02818}, 2021.

\bibitem{luccioni2024stable}
S.~Luccioni, C.~Akiki, M.~Mitchell, and Y.~Jernite, ``Stable bias: Evaluating societal representations in diffusion models,'' \emph{Advances in Neural Information Processing Systems}, vol.~36, 2024.

\bibitem{assran2022masked}
M.~Assran, M.~Caron, I.~Misra, P.~Bojanowski, F.~Bordes, P.~Vincent, A.~Joulin, M.~Rabbat, and N.~Ballas, ``Masked siamese networks for label-efficient learning,'' in \emph{European Conference on Computer Vision}.\hskip 1em plus 0.5em minus 0.4em\relax Springer, 2022, pp. 456--473.

\bibitem{he2022masked}
K.~He, X.~Chen, S.~Xie, Y.~Li, P.~Doll{\'a}r, and R.~Girshick, ``Masked autoencoders are scalable vision learners,'' in \emph{Proceedings of the IEEE/CVF conference on computer vision and pattern recognition}, 2022, pp. 16\,000--16\,009.

\bibitem{lin2014microsoft}
T.-Y. Lin, M.~Maire, S.~Belongie, J.~Hays, P.~Perona, D.~Ramanan, P.~Doll{\'a}r, and C.~L. Zitnick, ``Microsoft coco: Common objects in context,'' in \emph{Computer Vision--ECCV 2014: 13th European Conference, Zurich, Switzerland, September 6-12, 2014, Proceedings, Part V 13}.\hskip 1em plus 0.5em minus 0.4em\relax Springer, 2014, pp. 740--755.

\bibitem{young2014image}
P.~Young, A.~Lai, M.~Hodosh, and J.~Hockenmaier, ``From image descriptions to visual denotations: New similarity metrics for semantic inference over event descriptions,'' \emph{Transactions of the Association for Computational Linguistics}, vol.~2, pp. 67--78, 2014.

\bibitem{vaswani2017attention}
A.~Vaswani, ``Attention is all you need,'' \emph{Advances in Neural Information Processing Systems}, 2017.

\bibitem{zhai2023sigmoid}
X.~Zhai, B.~Mustafa, A.~Kolesnikov, and L.~Beyer, ``Sigmoid loss for language image pre-training,'' in \emph{Proceedings of the IEEE/CVF International Conference on Computer Vision}, 2023, pp. 11\,975--11\,986.

\bibitem{yang2022unified}
J.~Yang, C.~Li, P.~Zhang, B.~Xiao, C.~Liu, L.~Yuan, and J.~Gao, ``Unified contrastive learning in image-text-label space,'' in \emph{Proceedings of the IEEE/CVF Conference on Computer Vision and Pattern Recognition}, 2022, pp. 19\,163--19\,173.

\bibitem{seneviratne2015early}
S.~Seneviratne, A.~Seneviratne, M.~A. Kaafar, A.~Mahanti, and P.~Mohapatra, ``Early detection of spam mobile apps,'' in \emph{Proceedings of the 24th International Conference on World Wide Web}, 2015, pp. 949--959.

\bibitem{rajasegaran2019multi}
J.~Rajasegaran, N.~Karunanayake, A.~Gunathillake, S.~Seneviratne, and G.~Jourjon, ``A multi-modal neural embeddings approach for detecting mobile counterfeit apps,'' in \emph{The World Wide Web Conference}, 2019, pp. 3165--3171.

\bibitem{kingma2014adam}
D.~P. Kingma and J.~Ba, ``Adam: A method for stochastic optimization,'' \emph{arXiv preprint arXiv:1412.6980}, 2014.

\bibitem{loshchilov2016sgdr}
I.~Loshchilov and F.~Hutter, ``Sgdr: Stochastic gradient descent with warm restarts,'' \emph{arXiv preprint arXiv:1608.03983}, 2016.

\bibitem{he2016deep}
K.~He, X.~Zhang, S.~Ren, and J.~Sun, ``Deep residual learning for image recognition,'' in \emph{Proceedings of the IEEE conference on computer vision and pattern recognition}, 2016, pp. 770--778.

\bibitem{dosovitskiy2020image}
A.~Dosovitskiy, L.~Beyer, A.~Kolesnikov, D.~Weissenborn, X.~Zhai, T.~Unterthiner, M.~Dehghani, M.~Minderer, G.~Heigold, S.~Gelly \emph{et~al.}, ``An image is worth 16x16 words: Transformers for image recognition at scale,'' \emph{arXiv preprint arXiv:2010.11929}, 2020.

\bibitem{li2022blip}
J.~Li, D.~Li, C.~Xiong, and S.~Hoi, ``Blip: Bootstrapping language-image pre-training for unified vision-language understanding and generation,'' in \emph{International conference on machine learning}.\hskip 1em plus 0.5em minus 0.4em\relax PMLR, 2022, pp. 12\,888--12\,900.

\bibitem{vig_2019_multiscale}
\BIBentryALTinterwordspacing
J.~Vig, ``A multiscale visualization of attention in the transformer model,'' in \emph{Proceedings of the 57th Annual Meeting of the Association for Computational Linguistics: System Demonstrations}.\hskip 1em plus 0.5em minus 0.4em\relax Florence, Italy: Association for Computational Linguistics, Jul. 2019, pp. 37--42. [Online]. Available: \url{https://www.aclweb.org/anthology/P19-3007}
\BIBentrySTDinterwordspacing

\bibitem{notdisgisechil.}
\BIBentryALTinterwordspacing
G.~P.~C. Help, ``Manage target audience and app content settings.'' [Online]. Available: \url{https://support.google.com/googleplay/android-developer/answer/9867159?hl=en}
\BIBentrySTDinterwordspacing

\bibitem{denipitiyage2024detecting}
D.~Denipitiyage, B.~Silva, K.~Gunathilaka, S.~Seneviratne, A.~Mahanti, A.~Seneviratne, and S.~Chawla, ``Detecting and characterising mobile app metamorphosis in {Google} {Play} store,'' \emph{arXiv preprint arXiv:2407.14565}, 2024.

\bibitem{2023gplypolicyviolation}
``How we fought bad apps and bad actors in 2023,'' \url{https://security.googleblog.com/2024/04/how-we-fought-bad-apps-and-bad-actors-in-2023.html}, 2024, accessed: 2024-08-07.

\bibitem{karunanayake2020multi}
N.~Karunanayake, J.~Rajasegaran, A.~Gunathillake, S.~Seneviratne, and G.~Jourjon, ``A multi-modal neural embeddings approach for detecting mobile counterfeit apps: A case study on {Google} {Play} store,'' \emph{IEEE Transactions on Mobile Computing}, vol.~21, no.~1, pp. 16--30, 2020.

\bibitem{steinbock2024comparing}
M.~Steinb{\"o}ck, J.~Bleier, M.~Rainer, T.~Urban, C.~Utz, and M.~Lindorfer, ``Comparing apples to {Android}s: Discovery, retrieval, and matching of and {Android} apps for cross-platform analyses,'' in \emph{2024 IEEE/ACM 21st International Conference on Mining Software Repositories (MSR)}.\hskip 1em plus 0.5em minus 0.4em\relax IEEE, 2024, pp. 348--360.

\end{thebibliography}

\begin{IEEEbiography}[{\includegraphics[width=1in,height=1.25in,clip,keepaspectratio]{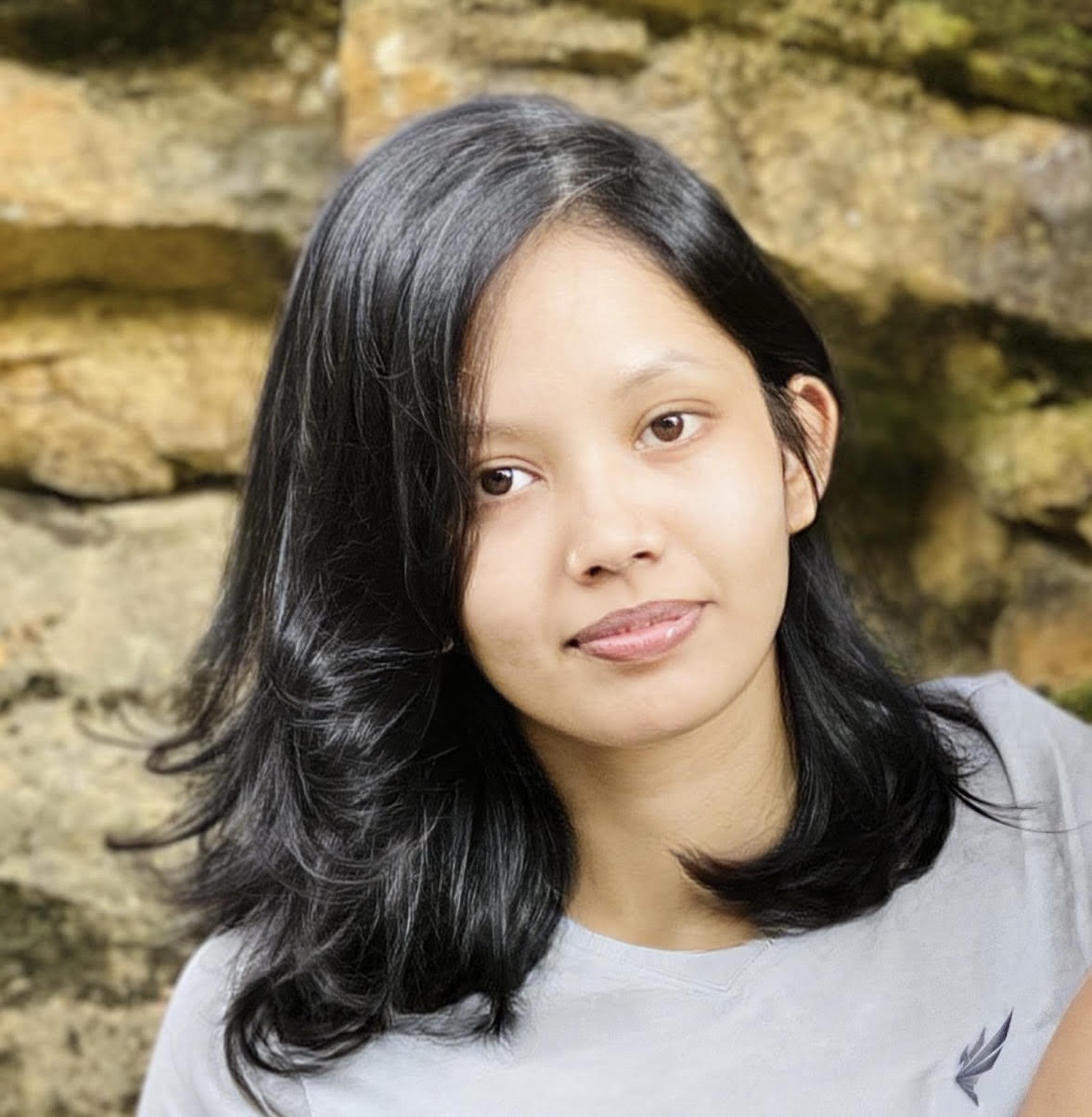}}]{Dishanika Denipitiyage}
	received her Bachelors degree in Electronic and Telecommunication Engineering from University of Moratuwa, Sri Lanka in 2020. She is currently working toward the PhD degree with the School of Computer Science, University of Sydney, Australia. She worked as a Senior software engineear at SenzMate (Pvt) Ltd, Sri Lanka in 2022 and as a Visiting Research Intern at Singapore University of Technology and Design (SUTD), Singapore in 2017. Her research interests include Self-Supervised Learning and Multi-Modal learning.
\end{IEEEbiography}

\vskip -3\baselineskip plus -1fil
\vspace{8mm}

\begin{IEEEbiography}[{\includegraphics[width=1in,height=1.5in,clip,keepaspectratio]{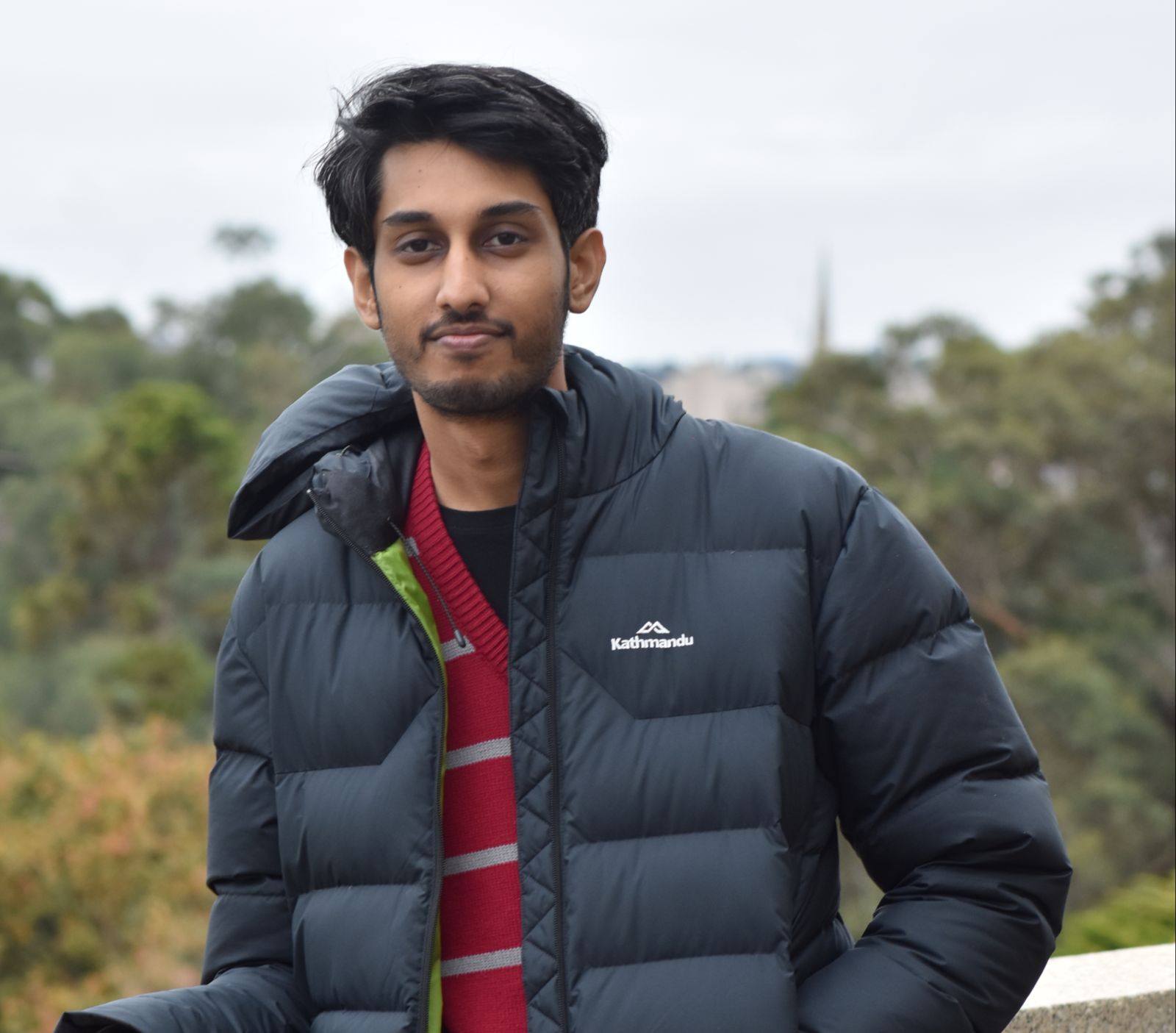}}]{Bhanuka Silva}
received his Bachelors degree in Electronic and Telecommunication Engineering (First Class Hons.) from University of Moratuwa, Sri Lanka in 2020. He also worked as a Visiting Research Intern at Data61-CSIRO, Brisbane in 2018 and is currently a doctoral student at the University of Sydney and his current research focuses on conducting privacy compliance checks in mobile app eco-systems by leveraging state-of-the art natural language processing techniques.
\end{IEEEbiography}

\vskip -3\baselineskip plus -1fil
\vspace{8mm}

\begin{IEEEbiography}[{\includegraphics[width=1in,height=1.25in,clip,keepaspectratio]{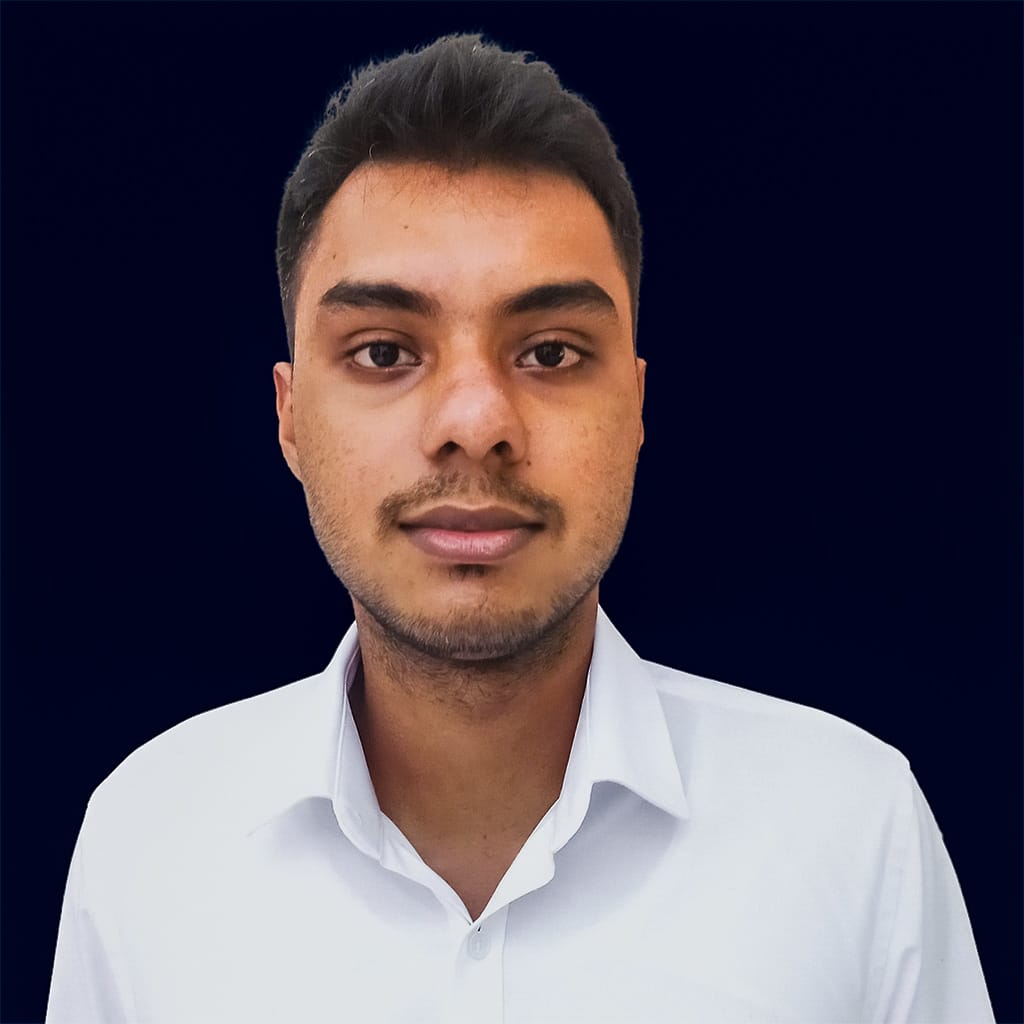}}]{Kavishka Gunathilaka}
is a B.Sc. Eng. undergraduate in the Department of Computer Science and Engineering of the University of Moratuwa, Sri Lanka. He also worked as a Research Affiliate at the University of Sydney in 2023. His primary research interests include machine learning, data 
science, and artificial intelligence.
\end{IEEEbiography}

\vskip -3\baselineskip plus -1fil
\vspace{8mm}

\begin{IEEEbiography}[{\includegraphics[width=1in,height=1.25in,clip,keepaspectratio]{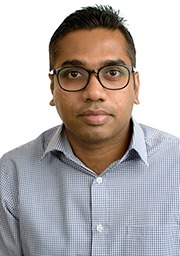}}]{Suranga Seneviratne}
is a Senior Lecturer in Security at the School of Computer Science, The University of Sydney. He received his Ph.D. from the University of New South Wales, Australia in 2015. His current research interests include privacy and security in mobile systems, AI applications in security, and behavior biometrics. Before moving into research, he worked nearly six years in the telecommunications industry in core network planning and operations. He received his bachelor degree from University of Moratuwa, Sri Lanka in 2005.
\end{IEEEbiography}
%\vspace{-4.2cm}

\vskip -2\baselineskip plus -1fil
\vspace{8mm}

\begin{IEEEbiography}[{\includegraphics[width=1in,height=1.25in,clip,keepaspectratio]{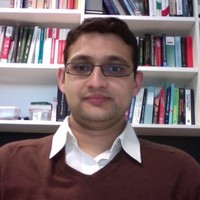}}]{Anirban Mahanti}'s technical expertise is at the intersection of computer networking, data science, and machine learning. Anirban earned his Ph.D. in Computer Science from the University of Saskatchewan, Canada, in 2003, his MSc in Computer Science in 1999, and his BE in Computer Science and Engineering from the Birla Institute of Technology, India, in 1993. He is currently an Honorary Senior Research Fellow at the University of Sydney.
\end{IEEEbiography}
\vskip -2\baselineskip plus -1fil
\begin{IEEEbiography}[{\includegraphics[width=1in,height=1.25in,clip,keepaspectratio]{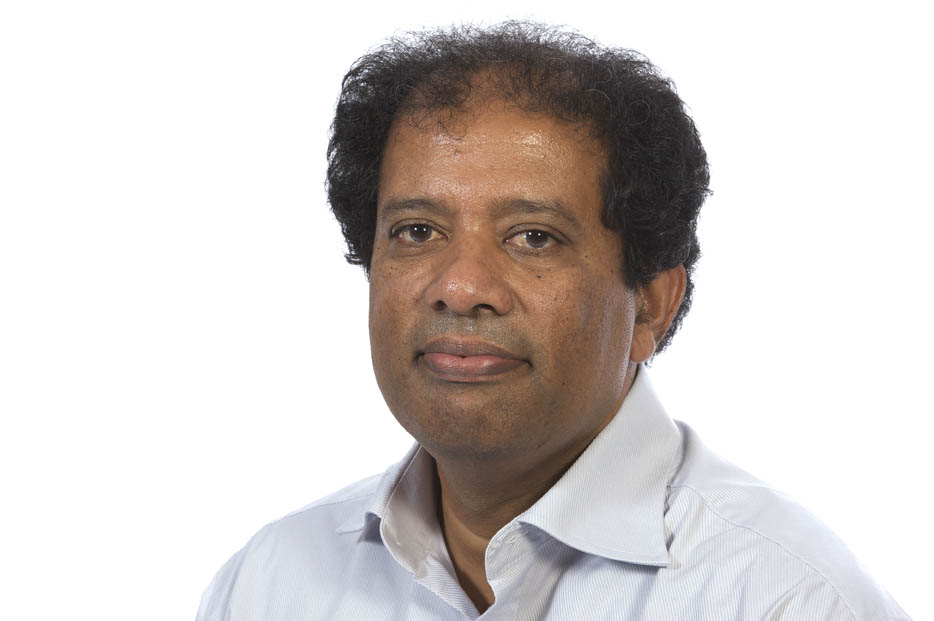}}]{Aruna Seneviratne}
(Senior Member, IEEE) is currently a foundation professor of telecommunications with the University of New South Wales, Sydney, Australia, where he holds the Mahanakorn chair of telecommunications. He was with a number of other universities in Australia, UK, and France, as well as industrial organizations, including Muirhead, Standard Telecommunication Labs, Avaya Labs, and Telecom Australia (Telstra). He held visiting appointments with INRIA, France. His research inter- ests include physical analytics, technologies that enable applications to interact intelligently and securely with their environment in real time. Recently, his team has been working on using these technologies in behavioural biometrics, optimizing the performance of wearables, and IoT system verification. He was the recipient of several fellowships, including one at British Telecom and one at Telecom Australia Research Labs.
\end{IEEEbiography}

\vskip -2\baselineskip plus -1fil
\vspace{8mm}

\begin{IEEEbiography}[{\includegraphics[width=1in,height=1.25in,clip,keepaspectratio]{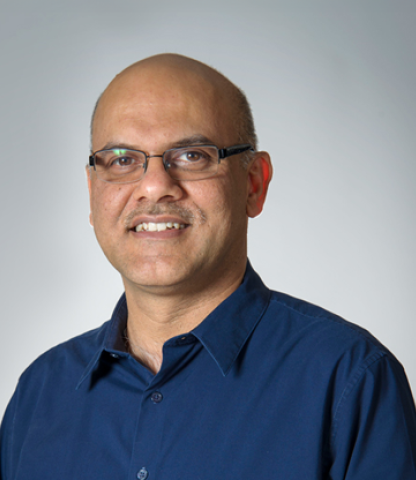}}]{Sanjay Chawla}
is Research Director of QCRI’s Data Analytics department. Prior to joining QCRI, Dr. Chawla was a Professor in the Faculty of Engineering and IT at the University of Sydney. From 2008-2011, he also served as the Head (Department Chair) of the School of Information Technologies. He was an academic visitor at Yahoo! Research in 2012. He received his PhD from the University of Tennessee (USA) in 1995. His research is in data mining and machine learning with a specialization in spatio-temporal data mining, outlier detection, class imbalanced classification, and adversarial learning. 
\end{IEEEbiography}

\end{document}